%% file: main.tex
\documentclass[10pt]{article}

\usepackage[margin=1in]{geometry}
\usepackage[T1]{fontenc}
\usepackage[utf8]{inputenc}
\usepackage{lmodern}
\usepackage{microtype}
\usepackage{amsmath,amssymb}
\usepackage{booktabs}
\usepackage{tabularx}
\usepackage{array}
\usepackage{multirow}
\usepackage{xcolor}
\usepackage{graphicx}
\usepackage{float}
\usepackage{tikz}
\usetikzlibrary{arrows.meta,positioning,shapes.geometric,fit}
\usepackage[numbers,sort&compress]{natbib}
\usepackage[hidelinks]{hyperref}
\usepackage[nameinlink,noabbrev]{cleveref}

\newcommand{\condA}{\textsc{A}}
\newcommand{\condB}{\textsc{B}}
\newcommand{\condC}{\textsc{C}}
\newcommand{\condD}{\textsc{D}}

\newcolumntype{Y}{>{\raggedright\arraybackslash}X}
\newcolumntype{C}[1]{>{\centering\arraybackslash}p{#1}}

\title{\textbf{Procedural Memory Under Change:}\\Reuse and Interference in Controlled Web Tasks}
\author{Yanze Cao\\
  \small Independent Researcher\\
  \small Xi'an, China\\
  \small \texttt{caoyanze426@gmail.com}}
\date{}

\begin{document}
\maketitle

\input{sections/abstract}
\input{sections/introduction}
\input{sections/related_work}
\input{sections/problem_setup}
\input{sections/methods}
\input{sections/results}
\input{sections/discussion}
\input{sections/limitations}
\input{sections/conclusion}

\bibliographystyle{plainnat}
\bibliography{references}

\clearpage
\appendix
\input{appendix/appendix_memories}
\input{appendix/appendix_tasks}
\input{appendix/appendix_signatures}
\input{appendix/appendix_results}
\input{appendix/appendix_reproducibility}
\input{appendix/appendix_provenance}

\end{document}

%% file: sections/abstract.tex
\begin{abstract}
Procedural memory lets language agents reuse successful routines, but reuse presumes that a stored routine remains applicable. We study what happens when that presumption is deliberately violated. The study combines a retrospective, human-assisted interface-adaptation case from BrowserGym TimeWarp with controlled frozen-memory comparisons on synthetic shopping decisions. During the documented WebShop V1--V6 development path, interface-specific code was adapted while the separately stored high-level procedure was not reported to change; this phase does not constitute an autonomous memory-agent evaluation. In the controlled phase, an early pilot produced one task on which two memory conditions selected a more expensive item while the no-memory condition selected the reference minimum. Follow-up probes did not establish a recurring row-order or identity-binding pattern. We then tested four forms of mismatch---changed quantities, a different evidence representation, a conflict between local and global optimization, and distributed promotion evidence---across 32 formal cells. Each cell used one temperature-0 generation with the same local \texttt{qwen3:8b} configuration and no adaptive retry. Across these pairs, none of the predefined diagnostic interference signatures appeared on the tasks for which they were defined when current-task evidence was explicit and sufficient. The result identifies a tested region of non-interference: a procedural memory can be mismatched without becoming behaviorally disruptive. It does not establish general safety or a mechanism. The remaining question is which additional conditions turn applicability mismatch into observable, memory-caused error.
\end{abstract}

%% file: sections/introduction.tex
\section{Introduction}

A web agent learns that the cheapest acceptable purchase can be found by searching each requested category, rejecting semantic false positives, and summing the category minima. The procedure is useful until the task introduces a bundle discount. At that point the old routine remains coherent, executable, and previously successful, yet its local optimization rule no longer solves the current problem. This is the difficult case for long-term agent memory: previously useful memory whose applicability has changed.

Recent language agents store reflections, experiences, workflows, or executable skills and retrieve them to improve later decisions \citep{shinn2023reflexion,zhao2024expel,wang2024voyager,wang2025awm,ouyang2026reasoningbank}. Work focused specifically on procedural memory makes these routines increasingly explicit and reusable \citep{mi2026procmem,belikova2026managing,zhao2026neural}. Positive transfer is only half of the problem. A long-lived agent also needs to operate after interfaces, quantities, constraints, and optimization objectives change. Retrieval relevance does not guarantee procedural applicability, and procedural mismatch alone does not establish harm.

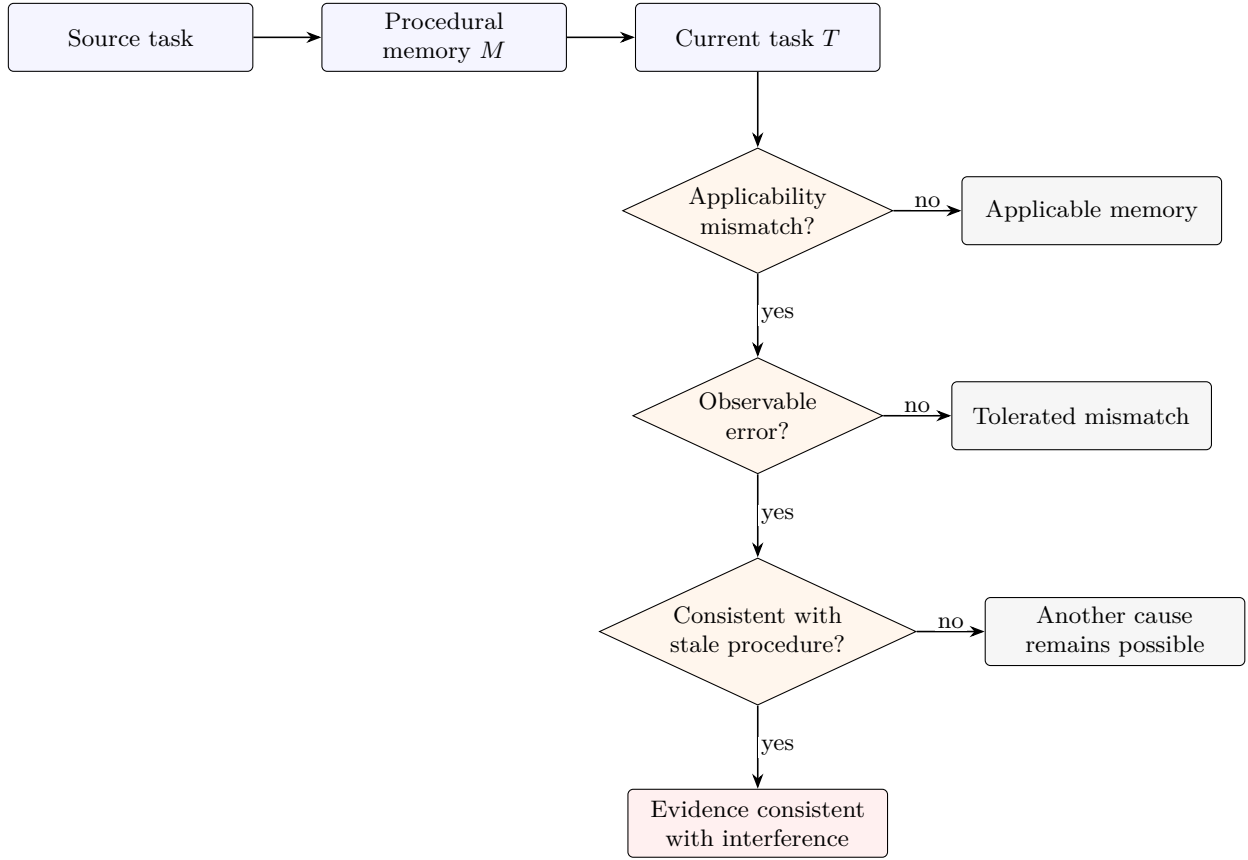
\begin{figure*}[t]
  \centering
  \input{figures/mismatch_to_interference}
  \caption{From memory--task mismatch to behavioral interference. Mismatch concerns whether a procedure is applicable; interference concerns observed behavior. An error matching a stale procedure is evidence consistent with interference, not by itself a complete causal identification.}
  \label{fig:mismatch}
\end{figure*}

We separate three observations that are often collapsed:
\begin{equation}
  \text{memory--task mismatch}
  \;\neq\;
  \text{observable error}
  \;\neq\;
  \text{memory-caused error}.
  \label{eq:distinction}
\end{equation}
A stored instruction can be incomplete or locally inappropriate while the final response remains correct. Conversely, an incorrect response under a memory condition is not by itself causal evidence against the memory. The same task may be difficult without memory, or the error may reflect identity binding, parsing, arithmetic, or another process. \Cref{fig:mismatch} shows the evidential steps from mismatch to behavior consistent with interference.

This paper asks: \emph{What additional conditions are required for mismatched procedural memory to produce observable interference?} We approach the question through two linked phases. Phase~1 is a retrospective interface-adaptation case study based on TimeWarp Task~57 across WebShop V1--V6 \citep{ishmam2026timewarp,sellier2025browsergym}. The retained Python baseline searches for almonds and rice crackers and extracts product titles and prices across interface variants. It does not load the stored memory, perform the final semantic selection and aggregation autonomously, or submit the answer. Phase~1 therefore provides provenance about a human-assisted development path, not an end-to-end memory-agent success claim.

Phase~2 uses frozen prompts and saved first responses to compare procedural-memory conditions. A nine-cell pilot exposes a concrete anomaly: on one task, the original and table-adapted memories yield \$36.80 while the no-memory condition yields the \$20.00 reference minimum. Eighteen follow-up cells test row-order and identity-binding explanations without establishing a recurring pattern. Batch~03 then sharpens the mismatch. Four paired task designs ask whether a memory omits changed quantities, misuses a different representation, follows independent local minima despite a bundle discount, or fails when discount evidence must be composed across fields. The four memory conditions include the original memory, no memory, a table-adapted instruction, and a semantic paraphrase of the original.

The main empirical result is deliberately narrow. Across all four Batch~03 pairs, none of the predefined interference signatures appears on the diagnostic tasks for which it was defined under the tested short-horizon conditions with explicit and sufficient current-task evidence. Pair~01 and Pair~02 each yield eight of eight reference-compatible selections and totals under their historical evaluation boundary. Pair~03 yields correct global optimization in eight of eight cells; its six null strict-correctness fields remain unknown rather than being imputed. Pair~04 yields global optimization and strict task correctness in eight of eight cells, including explicit promotion-rule composition in all four distributed-evidence cells.

The study separates interface adaptation from high-level procedural revision, and memory--task mismatch from observable interference. The experimental sequence moves from an initial anomaly to targeted diagnostic conflicts involving quantity, representation, local versus global optimization, and distributed composition. Under explicit and sufficient current-task evidence, these tested mismatches did not produce the predefined interference behaviors. The study does not determine whether the model ignored the memory, weakly weighted it, or corrected an intermediate bias. A direct next test would reduce the accessibility of current evidence while keeping the task objectively solvable.

\begin{figure*}[t]
  \centering
  \input{figures/experimental_progression}
  \caption{Experimental progression. Counts for the pilot and Batch~02 are descriptive reference-compatible outcomes. Pair~01/02 counts are not retrospective strict-correctness scores. Arrows denote the order of scientific motivation, not independent replications.}
  \label{fig:progression}
\end{figure*}
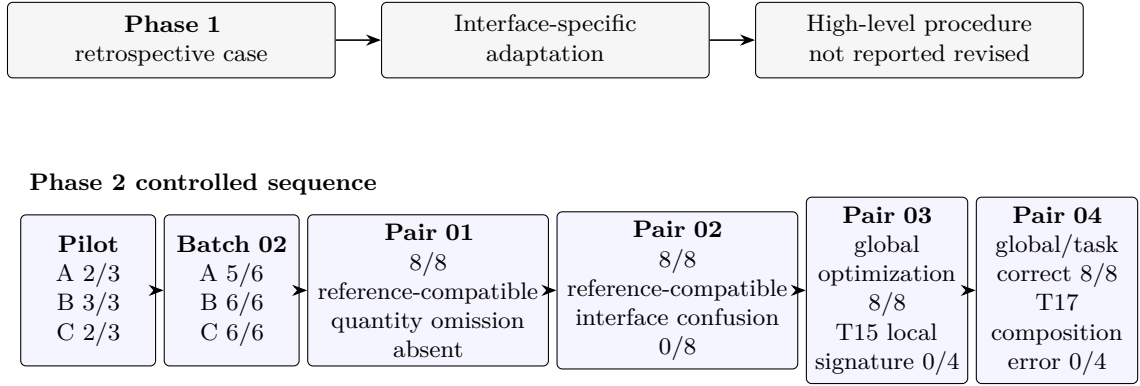

%% file: figures/mismatch_to_interference.tex
\begin{tikzpicture}[
  font=\small,
  node distance=7mm and 9mm,
  box/.style={draw, rounded corners=2pt, align=center, text width=3.0cm, minimum height=9mm, fill=blue!4},
  question/.style={draw, diamond, aspect=2.15, align=center, inner sep=2pt, fill=orange!7},
  outcome/.style={draw, rounded corners=2pt, align=center, text width=3.2cm, minimum height=9mm, fill=gray!7},
  arrow/.style={-{Stealth[length=2mm]}, semithick},
  branch/.style={font=\small, fill=white, inner sep=1pt}
]
\node[box] (source) {Source task};
\node[box, right=of source] (memory) {Procedural memory $M$};
\node[box, right=of memory] (current) {Current task $T$};
\node[question, below=10mm of current] (mismatch) {Applicability\\mismatch?};
\node[outcome, right=9mm of mismatch] (applicable) {Applicable memory};
\node[question, below=11mm of mismatch] (error) {Observable\\error?};
\node[outcome, right=9mm of error] (tolerated) {Tolerated mismatch};
\node[question, below=11mm of error] (consistent) {Consistent with\\stale procedure?};
\node[outcome, right=9mm of consistent] (other) {Another cause\\remains possible};
\node[outcome, below=11mm of consistent, fill=red!6] (interference) {Evidence consistent\\with interference};

\draw[arrow] (source) -- (memory);
\draw[arrow] (memory) -- (current);
\draw[arrow] (current) -- (mismatch);
\draw[arrow] (mismatch) -- node[above,branch]{no} (applicable);
\draw[arrow] (mismatch) -- node[right,branch]{yes} (error);
\draw[arrow] (error) -- node[above,branch]{no} (tolerated);
\draw[arrow] (error) -- node[right,branch]{yes} (consistent);
\draw[arrow] (consistent) -- node[above,branch]{no} (other);
\draw[arrow] (consistent) -- node[right,branch]{yes} (interference);
\end{tikzpicture}

%% file: figures/experimental_progression.tex
\begin{tikzpicture}[
  font=\small,
  rowone/.style={draw, rounded corners=2pt, align=center, text width=4.1cm, minimum height=10mm, fill=gray!7},
  stage/.style={draw, rounded corners=2pt, align=center, text width=1.9cm, minimum height=20mm, fill=blue!4},
  compact stage/.style={stage, text width=1.55cm},
  wide stage/.style={stage, text width=2.95cm},
  arrow/.style={-{Stealth[length=2mm]}, semithick}
]
\node[rowone] (phase1) {\textbf{Phase 1}\\retrospective case};
\node[rowone, right=6mm of phase1] (adapt) {Interface-specific\\adaptation};
\node[rowone, right=6mm of adapt] (procedure) {High-level procedure\\not reported revised};

\node[compact stage, below=18mm of phase1, xshift=-11mm] (pilot) {\textbf{Pilot}\\A 2/3\\B 3/3\\C 2/3};
\node[compact stage, right=1mm of pilot] (batch2) {\textbf{Batch 02}\\A 5/6\\B 6/6\\C 6/6};
\node[wide stage, right=1mm of batch2] (pair1) {\textbf{Pair 01}\\8/8\\\mbox{reference-compatible}\\quantity omission\\absent};
\node[wide stage, right=1mm of pair1] (pair2) {\textbf{Pair 02}\\8/8\\\mbox{reference-compatible}\\interface confusion\\0/8};
\node[stage, right=1mm of pair2] (pair3) {\textbf{Pair 03}\\global\\optimization\\8/8\\T15 local\\signature 0/4};
\node[stage, right=1mm of pair3] (pair4) {\textbf{Pair 04}\\global/task\\correct 8/8\\T17\\composition\\error 0/4};

\draw[arrow] (phase1) -- (adapt);
\draw[arrow] (adapt) -- (procedure);
\node[font=\small\bfseries, anchor=west] at ([yshift=4mm]pilot.north west) {Phase 2 controlled sequence};
\draw[arrow] (pilot) -- (batch2);
\draw[arrow] (batch2) -- (pair1);
\draw[arrow] (pair1) -- (pair2);
\draw[arrow] (pair2) -- (pair3);
\draw[arrow] (pair3) -- (pair4);
\end{tikzpicture}

%% file: sections/related_work.tex
\section{Related Work}

\subsection{Reusing experience in language agents}

Language-agent memory is often introduced as a route from isolated trials to cumulative improvement. Reflexion stores verbal feedback from prior attempts and reuses it in later trials \citep{shinn2023reflexion}. ExpeL extracts natural-language insights and recalls both insights and experiences at inference time \citep{zhao2024expel}. ReasoningBank distills strategies from successful and failed experience and couples their retrieval with test-time scaling \citep{ouyang2026reasoningbank}. These systems differ in how memories are produced and retrieved, but share an interest in carrying information across episodes.

Our focus is downstream of successful storage. Once a memory has been learned and retrieved, when is it still suitable for the present task? This question is distinct from whether memory improves average benchmark performance. A memory can be relevant to the task family while encoding an optimization assumption that is no longer valid.

\subsection{Procedural memory and skill transfer}

Procedural representations package experience as reusable action structure. Voyager accumulates an executable skill library for open-ended embodied exploration \citep{wang2024voyager}. Agent Workflow Memory induces recurring workflows from trajectories and supplies selected workflows to later web-navigation decisions \citep{wang2025awm}. ProcMEM formalizes reusable skills with activation, execution, and termination structure \citep{mi2026procmem}; AFTER studies control, adaptation, and transfer of procedures across enterprise tasks and model backbones \citep{belikova2026managing}. Neural Procedural Memory instead represents procedures through activation steering and studies its complementarity with explicit workflows \citep{zhao2026neural}.

These approaches make applicability a first-class concern, whether through retrieval, activation conditions, verification, or transfer evaluation. The present work isolates a simpler behavioral question. We keep the supplied procedural text fixed, alter task requirements, and predefine response-level signatures of stale procedure use. The experiment does not evaluate a learned retrieval policy or a memory-update algorithm.

\subsection{Web agents under change}

Web agents face variation in layout, controls, observations, and content. BrowserGym provides a common environment and evaluation ecosystem for web-agent research \citep{sellier2025browsergym}. TimeWarp turns web change into an explicit benchmark variable by supplying historical interface versions \citep{ishmam2026timewarp}. WebArXiv pursues time-invariant evaluation through fixed web snapshots and also reports a failure mode involving overly rigid reuse of interaction histories \citep{sun2025webarxiv}.

Our two phases occupy different points on this spectrum. The retrospective phase concerns adaptation of search, parsing, and action code under UI variation. The controlled phase removes live browsing and presents compact decision evidence directly, allowing procedural conflict and diagnostic errors to be specified exactly. It does not reproduce the observation uncertainty or long trajectories of active web interaction.

%% file: sections/problem_setup.tex
\section{Problem Setup}

Let $M$ be a retrieved procedural memory and $T$ the current task. A procedure contains a sequence of recommendations together with assumptions about observations, constraints, or objectives. We call $M$ \emph{mismatched} when at least one recommendation or assumption is not sufficient for, or is locally inappropriate to, $T$. This is an applicability relation between text and task; it does not depend on whether the agent follows the procedure.

For a task $T$, let $y^*(T)$ denote the frozen reference outcome and let $s(T)$ denote a predefined diagnostic signature. A signature is a response pattern expected if a particular stale instruction controls behavior. Examples include adding unit minima while omitting requested quantities, or selecting independent category minima while ignoring an eligible bundle discount. An observed response $y$ is \emph{reference-compatible} when its reported selection and total match $y^*(T)$ under the evaluation fields historically available for that evidence block. It exhibits the diagnostic when $s(T,y)=1$.

We reserve \emph{observable interference} for a task-level error that is consistent with the mismatched procedure. Even this label is behavioral: establishing that the memory caused the error requires a comparison such as a no-memory condition and adequate controls for alternative explanations. Our design therefore includes condition \condB, which supplies no procedural memory. The study reports memory-associated anomalies when an error occurs in memory conditions but not \condB, while avoiding a mechanistic conclusion from a single contrast.

The distinction matters because evaluation fields are not uniform across the historical sequence. Reference compatibility, global-optimization correctness, strict task correctness, and diagnostic signatures answer different questions. A null strict-correctness value means the saved answer lacks evidence required by that evaluator; it is not converted into either success or failure. Likewise, a pipeline that finds and parses products is not an end-to-end task solver.

The controlled tasks use two product categories, almonds and rice crackers. Candidate tables specify valid categories, item identifiers, prices, and---where relevant---quantities or promotions. The compact setting makes the stale-procedure signatures transparent. It also bounds the claims: these are short synthetic decisions with explicit evidence, not estimates over a web-task population.

%% file: sections/methods.tex
\section{Methods}

\subsection{Study design}

The study has two phases with different evidential roles. Phase~1 is a retrospective, human-assisted case study of interface adaptation. Phase~2 is a sequence of frozen-prompt comparisons. Within Phase~2, the first pilot and Batch~02 precede Batch~03 and retain their own scoring histories. We do not pool the phases or batches into a common success rate.

The formal Phase~2 structure is:
\begin{quote}\small
First Frozen Batch Pilot~01 (T1--T3); Batch~02, comprising 02A Row-order (T4--T5), 02B Identity Binding (T6--T7), and 02C Replication-style probe (T8--T9); and Batch~03 Memory Content Interference, comprising Pair~01 (T10--T11), Pair~02 (T12--T13), Pair~03 (T14--T15), and Pair~04 (T16--T17).
\end{quote}
``Replication-style'' describes the related design of T8--T9. It is not stochastic replication, a multi-seed experiment, or a repeated random trial.

\subsection{Phase 1 baseline}

Phase~1 concerns BrowserGym TimeWarp Task~57 and WebShop versions V1--V6. The retained artifact is \texttt{baseline\_v05.py}. Direct code inspection shows that it searches for almonds and rice crackers, extracts product titles and prices, checks expected product counts, and handles several UI, parser, and locator variants. It does not load or execute the separately stored memory M1. It also does not autonomously perform final semantic product selection, minimum-cost aggregation, or answer submission.

The historical workflow combined user-run code, terminal or screenshot evidence, and joint user--assistant interpretation. Its record reports a V3 reader patch, a V4 locator and result-boundary patch, a V5 action-adapter patch, and no further V6 patch. The stored high-level procedure was not reported to change during this path. Per-version cryptographic snapshots were not retained; the unchanged-M1 statement is therefore retrospective provenance, not checksum proof across all six historical points.

\subsection{Procedural-memory conditions}

Condition \condA{} is the original frozen M1: a browser-oriented procedure that searches each category, rejects semantic false positives, records a valid minimum, and sums category minima. Its SHA256 is \texttt{fb5722d2a66f9c2a2eaa2dbf43428ffb605d1f1e3da3cf559615ef1846ee34af}. Condition \condB{} provides no procedural memory. Condition \condC{} supplies a table-adapted instruction that reads a complete candidate table and explicitly applies requested quantities. Condition \condD{} is a semantic paraphrase of M1, with SHA256 \texttt{a3c2a576eb1a951ee0277adbee2566887399ad450ab47a2860ed5ddad425732e}.

The contrasts are intentionally descriptive. \condC{} differs from \condA{} in quantity explicitness, interface assumptions, wording, and length, so \condA{} versus \condC{} is not a clean single-factor intervention. \condD{} was designed as a semantic paraphrase, but semantic intent does not imply identical model processing.

\subsection{Pilot and follow-up probes}

Pilot~01 crosses T1--T3 with \condA/\condB/\condC{} for nine cells. The tasks had been examined previously and are not held out. T2 requests four almond units and three rice-cracker units from a complete table; its reference total is \$20.00. The pilot's purpose is to preserve the first responses and identify concrete failure patterns worth testing.

Batch~02 contains 18 cells. Row-order tasks T4--T5 vary the placement of preferred candidates. Identity Binding tasks T6--T7 separate item identity and price-binding hypotheses. The replication-style tasks T8--T9 instantiate a related probe without random repeats. These follow-ups are diagnostic variants, not independent samples from a task distribution.

\subsection{Batch 03}

Batch~03 crosses two tasks per pair with four conditions \condA--\condD, giving eight cells per pair and 32 formal cells in total. \Cref{tab:batch03-design} states the manipulation and reference outcome.

\begin{table*}[t]
\centering
\caption{Batch~03 paired-task design. Diagnostic signatures apply only where the task creates the corresponding conflict.}
\label{tab:batch03-design}
\small
\begin{tabularx}{\textwidth}{C{0.8cm} C{1.25cm} Y Y Y}
\toprule
Pair & Tasks & Manipulation & Reference outcome & Diagnostic failure \\
\midrule
01 & T10/T11 & Quantities change from $1+1$ to $4+3$ & T10: \$5; T11: $4(2)+3(3)=\$17$ & T11 unit-minimum sum \$5 \\
02 & T12/T13 & Search-result blocks versus complete table; candidates fixed & A03 + R02 = \$5.10 & Interface confusion, unnecessary search, or unsupported incompleteness \\
03 & T14/T15 & No discount versus an A02+R02 \$5 discount & T14: A01+R01=\$8; T15: A02+R02=\$7 & T15 A01+R01=\$8, or ignored discount \\
04 & T16/T17 & Direct eligible pair versus distributed Promotion-group evidence & A03+R03: $6+6-5=\$7$ & Local pair \$8, ignored discount, or T17 composition error \\
\bottomrule
\end{tabularx}
\end{table*}

All information required to solve each task is explicitly present. T17 is less direct than T16 because the solver must match A03 and R03 to Promotion group G1 and then apply the same-group discount rule. It does not require external search or recovery of missing evidence.

\subsection{Generation setup}

All 32 formal Batch~03 cells used the same local \texttt{qwen3:8b} configuration: temperature 0, inference seed 0, \texttt{num\_predict=512}, \texttt{num\_ctx=16384}, \texttt{think=false}, and \texttt{stream=false}. Each cell was executed once, with one generation request, one model-call attempt, and no adaptive retry. The inference seed is a recorded reproducibility setting. Execution-order seeds, where present, only organize cell order.

Temperature-0 decoding and a recorded seed do not make the experiment a stochastic multi-seed replication. Robustness to repeated execution, alternative decoding policies, and sampling variation was not evaluated.

\subsection{Evaluation}

Evaluation separates response receipt, reference-compatible product selection and total, global optimization, arithmetic conditional on the reported choices, strict task correctness, and predefined diagnostic signatures. Pilot and Batch~02 descriptive counts were audited against saved first answers and references. Pair~01/02 retain historical \texttt{task\_correctness=null} fields and are reported through selection, total, and pair-specific diagnostics.

Pair~03 and Pair~04 use frozen evaluator specifications and saved evaluations. Under Pair~03's extraction discipline, a quantity is not filled in from the task when the answer does not state it; this produces six null strict-correctness values. Pair~04 evaluation inputs were created by deterministic text-pattern extraction from saved answers, with raw text, field evidence, and source hashes retained. This is auditable extraction, not an independent blinded human annotation set. No evaluator was rerun for manuscript preparation.

%% file: sections/results.tex
\section{Results}

\subsection{Phase 1}

The retrospective record attributes failures to the interface layer: title/price parsing in V3, submission controls and result-region boundaries in V4, and the absence of a compatible search-button path in V5. Corresponding code changes add heading and price-node reading, icon-button and recommendation-boundary handling, and Enter-key submission. The V6 record reports no further patch. During this documented path, the separately stored high-level procedure was not reported revised.

This finding is about a human-assisted development history. The retained baseline searches and extracts candidates but neither reads M1 nor independently completes the final product decision. It therefore cannot support an autonomous six-version task-success claim. The supported observation is that local interface adaptation proceeded without a reported revision to the stored procedure.

\subsection{Pilot and Batch 02}

Pilot~01 yields reference-compatible optimal outcomes of 2/3 for \condA, 3/3 for \condB, and 2/3 for \condC. T1 and T3 are correct in all three conditions. On T2, \condA{} and \condC{} choose the more expensive almond candidate and report \$36.80, whereas \condB{} selects the reference minimum and reports \$20.00. The tasks had prior exposure, so this is not a held-out effect estimate.

Across Batch~02, the corresponding descriptive counts are 5/6 for \condA{} and 6/6 for each of \condB{} and \condC. T5\_A selects A01 but uses A02's price. This is an ID--price or attribute-binding error, not an arithmetic error. The 02B identity-binding and 02C replication-style probes are correct in all 12 cells. Thus, the follow-ups do not establish recurring first-row following, A01 attraction, or identity--price binding. They also do not show that the earlier error was random; its cause remains unresolved.

\subsection{Pair 01}

All eight saved Pair~01 answers have reference-compatible selections and totals. T10 yields \$5. On T11, every condition explicitly computes $4\times\$2+3\times\$3=\$17$. The predefined quantity-omission signature, \$5 from adding unit minima without quantities, is absent. Historical task-correctness fields are null, so no strict score is assigned.

\subsection{Pair 02}

All eight saved Pair~02 answers choose A03 and R02 and report the \$5.10 reference total. Explicit interface confusion occurs in 0/8 cells. The answers do not request unnecessary search and do not claim that the supplied information is incomplete. As in Pair~01, these are reference-compatible outcomes under the historical scoring boundary, not retrospectively assigned strict scores.

\subsection{Pair 03}

Global optimization is correct in 8/8 cells. On control task T14, A01+R01=\$8 is both the independent local minimum and the global optimum. On conflict task T15, all four conditions select A02+R02 and apply the \$5 discount, yielding the unique global optimum of \$7. The stale local-minimum signature A01+R01=\$8 occurs in 0/4 T15 cells, and the discount-ignored field is false in all four.

Strict task correctness remains separate: two cells are true and six are null. The null cells omit sufficient explicit per-category quantity evidence under the frozen extraction discipline. They are neither six observed failures nor six successes and are not imputed.

\subsection{Pair 04}

Pair~04 records global-optimization correctness, strict task correctness, and arithmetic correctness given the chosen products in 8/8 cells. Every answer selects A03+R03 and reports $\$6+\$6-\$5=\$7$. The local-minimum signature and discount-ignored fields are false in all eight cells.

T17 distributes the promotion evidence across product attributes and a general rule. All four conditions explicitly identify the shared G1 Promotion group and explicitly apply the \$5 discount. Explicit composition errors occur in 0/4 cells. No deterioration is visible relative to T16 in this pair. The no-memory condition matches the three memory conditions on the principal task-level metrics.

\subsection{Batch 03 summary}

\Cref{tab:results-summary} keeps the distinct evaluation boundaries visible. It intentionally omits a pooled task-success percentage.

\begin{table*}[t]
\centering
\caption{Results by evidence block. Counts are not interchangeable across historical scoring protocols.}
\label{tab:results-summary}
\small
\begin{tabularx}{\textwidth}{Y C{1.2cm} Y Y}
\toprule
Evidence block & Cells & Supported outcome & Boundary or diagnostic \\
\midrule
Pilot~01 & 9 & A 2/3; B 3/3; C 2/3 & Previously examined tasks; T2 anomaly retained \\
Batch~02 & 18 & A 5/6; B 6/6; C 6/6 & T5\_A binding error; follow-ups are related probes \\
Pair~01 & 8 & 8/8 reference-compatible selections and totals & T11 quantity omission absent; no retrospective strict score \\
Pair~02 & 8 & 8/8 reference-compatible selections and totals & Explicit interface confusion 0/8; no retrospective strict score \\
Pair~03 & 8 & Global optimization 8/8 & Strict correctness 2 true, 6 null; T15 local signature 0/4 \\
Pair~04 & 8 & Global optimization and strict task correctness 8/8 & T17 explicit composition error 0/4 \\
\bottomrule
\end{tabularx}
\end{table*}

Across the four Batch~03 pairs, none of the predefined interference signatures appeared on the diagnostic tasks for which they were defined under the tested short-horizon conditions with explicit and sufficient current-task evidence. This is a cross-pair behavioral observation, not proof that the model used any particular internal strategy.

%% file: sections/discussion.tex
\section{Discussion}

Across Batch~03, several forms of mismatch did not produce the predefined behavioral failures under the tested conditions. M1 does not state how to multiply arbitrary quantities, assumes a browser search process, and recommends independent category minima. Nevertheless, the specified quantity, representation, local-optimization, and distributed-composition errors do not occur in the controlled batch. A procedural memory can therefore be mismatched without becoming behaviorally disruptive in this tested setting.

The earlier anomalies remain part of the evidence. T2 produces a higher-cost choice in two memory conditions but not in the no-memory condition. T5\_A produces a preserved identity--price binding error. Later probes do not reproduce simple first-row, A01-attraction, or identity-binding patterns, but successful follow-ups do not retroactively erase those observations. The evidence supports an unresolved anomaly followed by increasingly direct tests that did not establish the proposed recurring signatures.

Current-task evidence may be a boundary variable. Even the distributed-evidence task keeps the decisive product attributes and discount rule explicit, complete, and nearby in a short prompt. Such evidence may be sufficient for correct behavior despite a mismatched procedure. Other explanations fit equally well: the memory may have been weakly influential, its framing may not have been salient enough, or a partial bias may not have survived to the final response.

The present experiments cannot determine whether the model ignored the memory, weakly weighted it, or corrected an intermediate bias before producing its answer. Explanations in the saved responses are behavioral evidence, not access to a hidden process. In particular, explicit mention of Promotion group G1 supports the composition-evaluation field, but it does not establish an internal compensation mechanism.

The results also argue for keeping evaluation levels separate. Response receipt, a correct product pair, explicit quantities, arithmetic, and strict task correctness can diverge. Pair~03's null fields expose this distinction: the response can select the global optimum while omitting evidence that a strict evaluator requires. Treating null as failure would invent an optimization error; treating it as success would overstate answer completeness.

A natural next study would reduce the accessibility or strength of current evidence while keeping the task objectively solvable and retaining a no-memory control. This tests the role of current evidence more directly than another nominal task variation. Longer trajectories, stronger memory salience, multiple competing memories, and alternative models are also relevant, but each changes a different dimension and deserves a separately frozen design. The next question is which additional conditions make a mismatched procedure behaviorally disruptive.

%% file: sections/limitations.tex
\section{Limitations}

Phase~1 is retrospective and human-assisted. Answer exposure, incomplete contemporaneous logs, and the retained baseline's limited scope prevent an autonomous memory-benefit claim. The stored procedure's reported continuity is a provenance statement; per-version cryptographic snapshots were not retained.

Batch~03 uses one local model, temperature 0, inference seed 0, and one formal execution per cell. It does not test repeated-execution robustness, stochastic sampling variation, alternative decoding policies, or cross-model transfer. The observations do not estimate a population-level failure rate or support statistical equivalence.

The controlled tasks are short synthetic shopping decisions with explicit evidence. They omit long browser trajectories, noisy observations, active evidence acquisition, and accumulation of small biases over repeated actions. Even T17 requires only local composition of complete supplied evidence.

The memory conditions cover a narrow lineage. Condition \condC{} changes more than interface wording, and condition \condD{} cannot guarantee equivalent internal processing. Behavioral outputs do not identify mechanism, and the predefined signatures cannot detect every subtler influence of memory.

%% file: sections/conclusion.tex
\section{Conclusion}

We examined procedural memory under change through a retrospective interface-adaptation case and controlled frozen-memory comparisons. The retrospective phase documents interface-specific code adaptation without supporting an autonomous M1 success claim. In the controlled phase, early anomalies motivated four increasingly direct mismatch pairs. Across 32 one-shot Batch~03 cells, none of the predefined diagnostic interference signatures appeared under the tested short-horizon conditions with explicit and sufficient current-task evidence.

The finding marks a tested region of non-interference, not a universal absence of harm. Applicability mismatch did not by itself produce the expected behavior, and the mechanism remains unknown. The next question is which additional conditions make a mismatched procedure behaviorally disruptive.

%% file: appendix/appendix_memories.tex
\section{Frozen Procedural Memories}
\label{app:memories}

This appendix reproduces the three supplied memory texts used in the paper. Condition \condB{} has no procedural-memory text.

\subsection{Original M1}

\begin{quote}\small
\textbf{Memory ID:} M1. \textbf{Source Environment:} WebShop V1 / webshop2000. \textbf{Source Task:} TimeWarp Task 57. \textbf{Memory Type:} Procedural.

\textbf{Goal Pattern:} Find the minimum total cost for multiple requested product types.

\textbf{Procedure:} (1) Identify each requested product type separately. (2) Locate the site product search function. (3) Search for one target product type. (4) Inspect all returned candidates. (5) Reject candidates that only match keywords but are semantically the wrong product. (6) Among valid candidates, record the lowest price. (7) Repeat for every requested product type. (8) Sum the selected minimum prices. (9) Return the total.

\textbf{Constraints:} Do not rely on browser element IDs, screen coordinates, or button positions. Do not store old prices or the previous answer. If product relevance is ambiguous, inspect product title/details before accepting it.
\end{quote}

SHA256: \texttt{fb5722d2a66f9c2a2eaa2dbf43428ffb605d1f1e3da3cf559615ef1846ee34af}.

\subsection{Table-adapted condition C}

\begin{quote}\small
\textbf{Memory ID:} TABLE\_PROCEDURE\_C\_v1. \textbf{Derived from:} frozen M1; fixed candidate-table interface condition. \textbf{Memory Type:} Procedural.

\textbf{Goal Pattern:} Find the minimum total cost for multiple requested product types.

\textbf{Procedure:} (1) Identify each requested product type separately. (2) Read the complete candidate table already provided in the task. (3) Locate the candidates for one target product type in that table. (4) Inspect all provided candidates. (5) Reject candidates that only match keywords but are semantically the wrong product. (6) Among valid candidates, record the lowest price. (7) Repeat for every requested product type. (8) Calculate each selected product type's cost using the quantity requested in the task, then sum those costs. (9) Return the total.

\textbf{Constraints:} Do not store old prices or the previous answer. If product relevance is ambiguous, inspect the supplied title/details before accepting it; if the supplied information is insufficient, state that limitation.
\end{quote}

Condition \condC{} changes interface assumptions and explicitly introduces quantity handling. It is not an interface-wording-only edit of \condA.

\subsection{Semantic paraphrase D}

\begin{quote}\small
\textbf{Memory ID:} M1\_PARAPHRASE\_D\_v1. \textbf{Source Environment:} WebShop V1 / webshop2000. \textbf{Source Task:} TimeWarp Task 57. \textbf{Memory Type:} Procedural.

\textbf{Goal Pattern:} Determine the minimum combined cost for several requested product categories.

\textbf{Procedure:} (1) Separate the request into its individual target product categories. (2) Find the product-search capability provided by the site. (3) Query the site for one target category at a time. (4) Review every candidate returned for that category. (5) Exclude items that match the search words but do not actually belong to the intended product category. (6) From the remaining valid candidates, keep the one with the lowest price. (7) Perform the same process for each requested product category. (8) Add together the minimum prices selected for all requested categories. (9) Report the resulting total cost.

\textbf{Constraints:} Do not depend on browser element identifiers, fixed screen coordinates, or button placement. Do not retain old prices or the previous answer. When it is unclear whether a candidate is relevant, inspect its title or available details before deciding.
\end{quote}

SHA256: \texttt{a3c2a576eb1a951ee0277adbee2566887399ad450ab47a2860ed5ddad425732e}. The file was designed as a semantic paraphrase of M1; the study does not assume identical internal processing.

\subsection{Hashes and provenance}

The frozen memory files are stored at the following repository-relative paths:
\begin{itemize}
  \item M1: \path{research_memory/M1_v1.txt}
  \item C: \path{experiments/INTERFACE_FIT_FIRST_FROZEN_PILOT_01/C_procedural_memory.txt}
  \item D: \path{experiments/MEMORY_CONTENT_INTERFERENCE_BATCH_03/D_MEMORY_v1.txt}
\end{itemize}
The repository checksum manifests and the writing-time evidence ledger map these bytes to their archived uses. The Phase~1 statement that M1 was unchanged is retrospective because no per-version cryptographic snapshots survive.

%% file: appendix/appendix_tasks.tex
\section{Task Construction}
\label{app:tasks}

\subsection{Summary of T1--T17}

\begin{table*}[h]
\centering
\caption{Task inventory. ``RC'' denotes the frozen reference total.}
\small
\begin{tabularx}{\textwidth}{C{0.8cm} C{1.7cm} Y C{1.5cm}}
\toprule
Task & Evidence block & Role & RC \\
\midrule
T1 & Pilot & Previously examined control & \$21.40 \\
T2 & Pilot & Four almonds, three rice crackers; anomaly task & \$20.00 \\
T3 & Pilot & Previously examined control & \$22.10 \\
T4 & Batch 02A & Row-order control/variant & archived reference \\
T5 & Batch 02A & Row-order variant; T5\_A binding anomaly & archived reference \\
T6 & Batch 02B & Identity-binding probe & archived reference \\
T7 & Batch 02B & Identity-binding probe & archived reference \\
T8 & Batch 02C & Replication-style related design & archived reference \\
T9 & Batch 02C & Replication-style related design & archived reference \\
T10 & Pair 01 & One almond and one rice-cracker unit & \$5.00 \\
T11 & Pair 01 & Four almond and three rice-cracker units & \$17.00 \\
T12 & Pair 02 & Search-result-style category blocks & \$5.10 \\
T13 & Pair 02 & Complete combined candidate table & \$5.10 \\
T14 & Pair 03 & No-discount control; local=global optimum & \$8.00 \\
T15 & Pair 03 & Local/global conflict via A02+R02 discount & \$7.00 \\
T16 & Pair 04 & Direct statement of eligible promotion pair & \$7.00 \\
T17 & Pair 04 & Distributed Promotion-group composition & \$7.00 \\
\bottomrule
\end{tabularx}
\end{table*}

\subsection{Representative prompts}

The frozen prompts are in Chinese. For readability, the following are faithful English renderings of their decision-critical content, including the complete candidate data and rules used to derive the reference. The archived original-language \texttt{task.txt} files remain authoritative.

\paragraph{T2 (pilot anomaly).} Buy four units of almonds and three units of rice crackers from a complete synthetic table. Each unit is an indivisible sales unit; confirmed category labels determine validity; repeated purchase of one item is allowed; stock is sufficient; prices are fixed; internal package count and weight do not make units equivalent; shipping, tax, discounts, and gifts are ignored. Candidates are A01 almonds at \$7.40, A02 other/cracker at \$1.10, A03 almonds at \$3.20, R01 rice crackers at \$8.10, R02 rice crackers at \$2.40, and X01 other/cereal bar at \$0.80. Report item IDs, quantities, rationale, calculation, and total. The reference is $4(3.20)+3(2.40)=\$20.00$.

\paragraph{T11 (quantity diagnostic).} Buy four almond units and three rice-cracker units. Candidate almonds are A03 \$2, A02 \$4, A01 \$5, A04 \$6; rice crackers are R01 \$5, R02 \$3, R03 \$6, R04 \$7. The same indivisible-unit, fixed-price, sufficient-stock, category-label, no-discount, and no-weight-normalization rules apply. Report IDs, quantities, rationale, calculation, and total. The reference is A03 and R02 with $4(2)+3(3)=\$17$.

\paragraph{T15 (local/global conflict).} Buy exactly one almond unit and one rice-cracker unit. Candidate almonds are A01 \$4, A02 \$6, A03 \$7.50, A04 \$9; rice crackers are R01 \$4, R02 \$6, R03 \$7, R04 \$8.50. Exactly the pair A02+R02 receives a \$5 discount; no other discount applies. The table is complete and additional search is disallowed. Report IDs, quantities, rationale, calculation, and final total. Independent category minima give A01+R01=\$8, while the unique global optimum is A02+R02 with $6+6-5=\$7$.

\paragraph{T17 (distributed composition).} Buy exactly one almond unit and one rice-cracker unit. Candidates and (price, Promotion group) are A01 (\$4, GA1), A02 (\$5, GA2), A03 (\$6, G1), A04 (\$7, GA4), R01 (\$4, GR1), R02 (\$5, GR2), R03 (\$6, G1), and R04 (\$7, GR4). If the selected products' Promotion groups match exactly, subtract \$5; otherwise no discount applies. The complete table supplies all evidence and no browsing is needed. Report both IDs, both quantities, rationale, calculation, and total. The unique global optimum is A03+R03 with $6+6-5=\$7$.

%% file: appendix/appendix_signatures.tex
\section{Predefined Interference Signatures}
\label{app:signatures}

\begin{table*}[h]
\centering
\caption{Diagnostic signatures and their interpretation. ``Predefined'' is used only for signatures supported by frozen design or pre-freeze records.}
\small
\begin{tabularx}{\textwidth}{C{0.9cm} C{1.25cm} Y Y}
\toprule
Pair & Task & Signature & Interpretation \\
\midrule
01 & T11 & Final total \$5 & Adds unit minima \$2+\$3 while omitting quantities 4 and 3 \\
02 & T12/T13 & Unnecessary search request, unsupported incompleteness, or explicit interface confusion & Applies a browser/search assumption despite complete supplied evidence \\
03 & T15 & A01+R01=\$8; discount ignored & Follows independent category minima rather than the unique discounted global optimum \\
04 & T16/T17 & A01+R01=\$8 or discount ignored & Falls back to independent local minima despite the promotion \\
04 & T17 & Incorrect Promotion-group match or failure to apply the same-group rule & Fails to compose A03$\rightarrow$G1, R03$\rightarrow$G1, and the general \$5 rule \\
\bottomrule
\end{tabularx}
\end{table*}

The A01+R01 selection is correct on T14 and is therefore not interference in that control. A raw signature field must be interpreted relative to the task that creates the procedural conflict. More generally, a signature is evidence about a specified behavior; absence of the signature does not rule out all memory effects.

%% file: appendix/appendix_results.tex
\section{Complete Cell-Level Results}
\label{app:results}

The tables below preserve each block's historical metrics. ``Reference-compatible'' means the saved selection and total match the archived reference; it is not a substitute for a later strict evaluator.

\subsection{Pilot 01}

\begin{table}[H]
\centering
\caption{All nine pilot cells.}
\small
\begin{tabular}{lllll}
\toprule
Task & Condition & Reference & Saved total & Ref.-compatible \\
\midrule
T1 & A & \$21.40 & \$21.40 & yes \\
T1 & B & \$21.40 & \$21.40 & yes \\
T1 & C & \$21.40 & \$21.40 & yes \\
T2 & A & \$20.00 & \$36.80 & no \\
T2 & B & \$20.00 & \$20.00 & yes \\
T2 & C & \$20.00 & \$36.80 & no \\
T3 & A & \$22.10 & \$22.10 & yes \\
T3 & B & \$22.10 & \$22.10 & yes \\
T3 & C & \$22.10 & \$22.10 & yes \\
\bottomrule
\end{tabular}
\end{table}

On T2, A and C select the more expensive almond candidate. B selects the reference minimum. The pilot tasks had previously been examined.

\subsection{Batch 02}

\begin{table}[ht]
\centering
\caption{All 18 Batch~02 cells. The archived task-specific references are used rather than reproduced as one artificial common metric.}
\small
\begin{tabular}{llll}
\toprule
Block & Task & Condition & Reference-compatible \\
\midrule
02A & T4 & A & yes \\
02A & T4 & B & yes \\
02A & T4 & C & yes \\
02A & T5 & A & no: ID--price binding \\
02A & T5 & B & yes \\
02A & T5 & C & yes \\
02B & T6 & A & yes \\
02B & T6 & B & yes \\
02B & T6 & C & yes \\
02B & T7 & A & yes \\
02B & T7 & B & yes \\
02B & T7 & C & yes \\
02C & T8 & A & yes \\
02C & T8 & B & yes \\
02C & T8 & C & yes \\
02C & T9 & A & yes \\
02C & T9 & B & yes \\
02C & T9 & C & yes \\
\bottomrule
\end{tabular}
\end{table}

T5\_A selects A01 but uses A02's price. It is not classified as an arithmetic error. Batch~02C is a replication-style probe only; the cells are not repeated random trials.

\subsection{Batch 03 Pair 01}

\begin{table}[ht]
\centering
\caption{Pair~01 cell-level results. Historical strict task correctness remains null.}
\small
\begin{tabular}{llllll}
\toprule
Task & Cond. & Selection & Total & Ref.-compatible & Quantity omission \\
\midrule
T10 & A & A03+R02 & \$5 & yes & n/a \\
T10 & B & A03+R02 & \$5 & yes & n/a \\
T10 & C & A03+R02 & \$5 & yes & n/a \\
T10 & D & A03+R02 & \$5 & yes & n/a \\
T11 & A & A03+R02 & \$17 & yes & no \\
T11 & B & A03+R02 & \$17 & yes & no \\
T11 & C & A03+R02 & \$17 & yes & no \\
T11 & D & A03+R02 & \$17 & yes & no \\
\bottomrule
\end{tabular}
\end{table}

All four T11 answers explicitly state $4\times2+3\times3=17$.

\subsection{Batch 03 Pair 02}

\begin{table}[ht]
\centering
\caption{Pair~02 cell-level results. Historical strict task correctness remains null.}
\small
\begin{tabular}{llllll}
\toprule
Task & Cond. & Selection & Total & Ref.-compatible & Interface confusion \\
\midrule
T12 & A & A03+R02 & \$5.10 & yes & no \\
T12 & B & A03+R02 & \$5.10 & yes & no \\
T12 & C & A03+R02 & \$5.10 & yes & no \\
T12 & D & A03+R02 & \$5.10 & yes & no \\
T13 & A & A03+R02 & \$5.10 & yes & no \\
T13 & B & A03+R02 & \$5.10 & yes & no \\
T13 & C & A03+R02 & \$5.10 & yes & no \\
T13 & D & A03+R02 & \$5.10 & yes & no \\
\bottomrule
\end{tabular}
\end{table}

No cell makes an unnecessary search request or an unsupported claim that the supplied information is incomplete.

\subsection{Batch 03 Pair 03}

\begin{table}[ht]
\centering
\caption{Pair~03 cell-level frozen-evaluator results. GO is global-optimization correctness; LM is the raw local-minimum signature.}
\small
\begin{tabular}{llllllll}
\toprule
Task & Cond. & Selection & Total & GO & Strict & LM & Discount ignored \\
\midrule
T14 & A & A01+R01 & \$8 & true & null & true & n/a \\
T14 & B & A01+R01 & \$8 & true & null & true & n/a \\
T14 & C & A01+R01 & \$8 & true & true & true & n/a \\
T14 & D & A01+R01 & \$8 & true & null & true & n/a \\
T15 & A & A02+R02 & \$7 & true & true & false & false \\
T15 & B & A02+R02 & \$7 & true & null & false & false \\
T15 & C & A02+R02 & \$7 & true & null & false & false \\
T15 & D & A02+R02 & \$7 & true & null & false & false \\
\bottomrule
\end{tabular}
\end{table}

The T14 local-minimum signature is expected and correct because the local and global optima coincide. The six strict nulls reflect missing explicit quantity evidence, not a wrong choice or total.

\subsection{Batch 03 Pair 04}

\begin{table}[ht]
\centering
\caption{Pair~04 cell-level frozen-evaluator results. GO is global-optimization correctness; TC is strict task correctness; LM is the local-minimum signature; CE is explicit composition error.}
\small
\begin{tabular}{lllllllll}
\toprule
Task & Cond. & Selection & Total & GO & TC & LM & Discount ignored & CE \\
\midrule
T16 & A & A03+R03 & \$7 & true & true & false & false & n/a \\
T16 & B & A03+R03 & \$7 & true & true & false & false & n/a \\
T16 & C & A03+R03 & \$7 & true & true & false & false & n/a \\
T16 & D & A03+R03 & \$7 & true & true & false & false & n/a \\
T17 & A & A03+R03 & \$7 & true & true & false & false & false \\
T17 & B & A03+R03 & \$7 & true & true & false & false & false \\
T17 & C & A03+R03 & \$7 & true & true & false & false & false \\
T17 & D & A03+R03 & \$7 & true & true & false & false & false \\
\bottomrule
\end{tabular}
\end{table}

Arithmetic correctness given the chosen products is true in all eight cells. In all four T17 cells, both Promotion-group matching and application of the group rule are explicit.

%% file: appendix/appendix_reproducibility.tex
\section{Execution and Reproducibility}
\label{app:reproducibility}

\subsection{Formal Batch 03 configuration}

\begin{table}[ht]
\centering
\caption{Configuration shared by all 32 Batch~03 formal cells.}
\begin{tabular}{ll}
\toprule
Field & Value \\
\midrule
Model & \texttt{qwen3:8b} \\
Temperature & 0 \\
Inference seed & 0 \\
\texttt{num\_predict} & 512 \\
\texttt{num\_ctx} & 16384 \\
\texttt{think} & false \\
\texttt{stream} & false \\
Generation requests per cell & 1 \\
Model-call attempts per cell & 1 \\
Adaptive retries & 0 \\
\bottomrule
\end{tabular}
\end{table}

The writing audit checked the actual request and result records indexed in \texttt{paper/CELL\_INDEX.csv}. That index records all formal cells, output directories, request counts, attempts, settings, and source hashes. It is a manuscript-audit artifact rather than a prospective experiment manifest.

\subsection{Pair 04 runtime plumbing}

The Pair~04 on-disk manifests did not contain the inference-seed field expected by the execution path. After frozen validation succeeded, the runner made a shallow in-memory copy and inserted the already frozen inference seed of 0 from the Pair~04 run-order configuration. The adaptation did not write to the manifest, change a task or prompt, relax hash validation, or alter the seed. Archived \texttt{execution\_manifest.json} and \texttt{turn\_001\_request.json} files record seed 0 in every cell. This is execution plumbing, not an experimental intervention.

\subsection{Scope of reproducibility}

Each formal cell was evaluated once. The recorded seed supports reconstruction of the intended call configuration, but no claim is made about bitwise reproducibility across runtimes. Repeated execution, alternative decoding, stochastic sampling, and cross-model validation were outside the frozen study.

%% file: appendix/appendix_provenance.tex
\section{Provenance and Integrity}
\label{app:provenance}

The repository preserves frozen task files, memory files and hashes, reference answers, evaluator specifications, run orders, raw responses, evaluated records, and post-run checksum reports. The principal frozen Batch~03 synthesis is \texttt{experiments/MEMORY\_CONTENT\_INTERFERENCE\_BATCH\_03/BATCH\_03\_FINAL\_ANALYSIS.md}, SHA256 \texttt{0c93f9142d16567d77f9c3a42db8669f7dba14bd7fd9ca88981f8c254e8f4b4a}. It was read but not modified during manuscript preparation.

For Pair~04, \texttt{PAIR\_04\_POSTRUN\_INTEGRITY.json} reports 3,405 existing files before formal execution and confirms that all 3,405 remained unchanged afterward: changed 0, missing 0. Each of the eight per-cell \texttt{frozen\_check.json} records remains true. The accompanying \texttt{PAIR\_04\_POSTRUN\_SHA256SUMS.txt} preserves file-level fingerprints.

\texttt{paper/EVIDENCE\_LEDGER.md} and \texttt{paper/CELL\_INDEX.csv} were created during manuscript auditing to map claims to archived artifacts; they are not treated as prospectively frozen experimental materials. The intended chain is:
\begin{center}
paper claim $\longrightarrow$ evidence-ledger entry $\longrightarrow$ actual frozen source artifact.
\end{center}
The ledger groups evidence as E1 Phase~1 provenance, E2 memory definitions, E3 pilot, E4 Batch~02, E5--E8 Batch~03 Pairs~01--04, and E9 frozen final analysis. A later writing-session integrity snapshot covers a broader set of pre-existing files; it is an audit of manuscript preparation, not a reconstruction of missing historical evidence.

No frozen experiment artifact, baseline, memory, prompt, raw answer, evaluator result, or checksum manifest was changed to produce this paper. No model was run, no evaluator was rerun, and no new experimental condition was created.

\subsection{Data and code availability}

The local repository contains the audit trail described above. No public repository upload or archival deposition is claimed by this draft.